\pdfoutput=1

\documentclass[11pt]{article}
\usepackage{amsmath}
\usepackage{xspace}
\usepackage{amsfonts}
\usepackage{subcaption}
\usepackage{bm}
\usepackage{multirow}
\usepackage{booktabs} 
\usepackage{IEEEtrantools}
\usepackage{pifont}
\usepackage{color, colortbl}
\usepackage[inline]{enumitem}
\usepackage{listings}
\usepackage{ulem}
\usepackage{amssymb}
\usepackage{authblk}

\usepackage{amsmath,amssymb}
\usepackage{listings}
\usepackage[most]{tcolorbox}
\tcbuselibrary{skins,breakable}
\usepackage[preprint]{acl}

\usepackage{times}
\usepackage{latexsym}

\usepackage[T1]{fontenc}

\usepackage[utf8]{inputenc}

\usepackage{microtype}

\usepackage{inconsolata}
\usepackage{makecell}
\usepackage{url}
\usepackage{xurl}
\usepackage[hang,flushmargin]{footmisc}
\usepackage{hyperref}

\usepackage{graphicx}
\definecolor{Highlight}{rgb}{0.89,0.89,0.94}

\newcommand{\mymodel}{\textsc{DAM}\xspace}

\title{Beyond Heuristics: A Decision-Theoretic Framework for Agent Memory Management}
\author{\bf Changzhi Sun}
\author{\bf Xiangyu Chen}
\author{\bf Jixiang Luo}
\author{\bf Dell Zhang\thanks{Corresponding authors.}}
\author{\bf Xuelong Li\protect\footnotemark[1]}
\affil{Institute of Artificial Intelligence (TeleAI), China Telecom}
\affil[ ]{\tt \{dell.z, xuelong\_li\}@ieee.org}

\begin{document}

\maketitle

\begin{abstract}

External memory is a key component of modern large language model (LLM) systems, enabling long-term interaction and personalization.
Despite its importance, memory management is still largely driven by hand-designed heuristics, offering little insight into the long-term and uncertain consequences of memory decisions.
In practice, choices about what to read or write shape future retrieval and downstream behavior in ways that are difficult to anticipate.
We argue that memory management should be viewed as a sequential decision-making problem under uncertainty, where the utility of memory is delayed and dependent on future interactions.
To this end, we propose \mymodel (Decision-theoretic Agent Memory), a decision-theoretic framework that decomposes memory management into immediate information access and hierarchical storage maintenance.
Within this architecture, candidate operations are evaluated via value functions and uncertainty estimators, enabling an aggregate policy to arbitrate decisions based on estimated long-term utility and risk.
Our contribution is not a new algorithm, but a principled reframing that clarifies the limitations of heuristic approaches and provides a foundation for future research on uncertainty-aware memory systems.

\end{abstract}

\section{Introduction}
\label{sec:intro}

Large language models (LLMs) are increasingly augmented with external memory to support long-term interaction, personalization, and continual use~\cite{zhangConversationalAgentsRAG2025,hu2025memory}.
For example, conversational agents must persist user preferences and past goals across sessions, while task-oriented assistants rely on memory to track intermediate results and constraints over extended workflows.
By persisting information beyond a single prompt, a memory layer like TeleMem\footnote{\url{https://github.com/TeleAI-UAGI/telemem}} enables LLMs to operate in realistic, long-lived settings and has become a foundational component of modern deployed systems~\cite{chhikara2025mem0,li2025memos}.

In current systems, however, memory management is still largely governed by hand-designed heuristics.
Common practices include storing interactions verbatim, periodically appending summaries to context~\cite{latimer2025hindsight,chen2025moom,fang2025lightmem}, or pruning entries based on simple recency or similarity thresholds~\cite{zhong2024memorybank,rasmussen2025zep}.
While often easy to deploy, these static strategies obscure the fundamental complexity of memory operations: determining why specific information should be retained, when it requires revision, and under what exact conditions it should be forgotten.
More importantly, these rigid heuristics lack the adaptability required to handle shifting user goals or evolving tasks, often leading to brittle system performance where critical context is lost or irrelevant noise accumulates over time.

A key difficulty is that the utility of memory is rarely immediate.
Information that appears irrelevant in the current interaction, such as a user’s long-term preference or a temporarily abandoned subgoal, may become highly valuable in future contexts~\cite{lampinen2025latent,pritzel2017neural,ritter2018been}.
Conversely, details that seem useful at present may prove redundant or even harmful if retrieved inappropriately later~\cite{parisotto2017neural}.
For instance, indiscriminately storing transient utterances can inject noise into long-term memory, degrading future retrieval quality~\cite{graves2014neural,santoro2016meta}, whereas prematurely deleting information based on short-term inactivity may irreversibly remove knowledge essential for future tasks~\cite{graves2016hybrid}.
Consequently, memory decisions influence model behavior in ways that are delayed, uncertain, and difficult to anticipate.

These observations suggest that memory management is fundamentally a sequential decision problem.
Decisions about what to remember, revise, or forget shape the memory state available in subsequent interactions, which in turn constrains future model outputs~\cite{bellman1966dynamic,howard1960dynamic}.
Crucially, the consequences of these decisions depend on future interactions that are unknown at decision time~\cite{puterman2014markov,bertsekas1997nonlinear}.
This combination of delayed consequences and uncertainty aligns memory management with the broader class of sequential decision-making problems under uncertainty~\cite{bellman1966dynamic,powell2022sequential}.
From this perspective, the central challenge shifts from merely applying static rules to optimizing a composite action space, balancing immediate information access (\emph{reading}) against the long-term maintenance of the memory store (\emph{writing}), in anticipation of uncertain future utility.

Building on this sequential decision perspective~\cite{powell2022sequential}, we propose \mymodel (Decision-theoretic Agent Memory), a decision-theoretic framework for memory management in large language models.
Rather than treating memory as a passive buffer or relying on ad-hoc heuristics, \mymodel explicitly formulates memory management as a structured sequential decision problem, in which the system must continuously arbitrate between accessing information to support the current context and modifying the memory store to better serve future interactions.
A central challenge in this setting is the presence of delayed utility and significant uncertainty: memory operations may incur immediate costs while yielding benefits only much later, and the long-term impact of a write decision is often uncertain at the time it is made.

To address these challenges, we introduce a modular decision architecture that decomposes memory management into two complementary components: a \textbf{Read Policy} responsible for selecting information with immediate contextual relevance, and a hierarchical \textbf{Write Policy} that governs long-term storage maintenance.
Within the write mechanism, independent sub-policies first propose candidate memory operations (e.g., add or delete), each accompanied by an explicit \textbf{Value Function} ($V$) that estimates the expected long-term benefit of the operation, and an \textbf{Uncertainty Estimator} ($\Sigma$) that quantifies the associated risk.
These proposals are then evaluated by an internal \textbf{Aggregate Policy}, which integrates value and uncertainty signals to arbitrate among competing actions and execute coherent, system-level memory updates.
Importantly, our contribution is not a single learning rule or optimization algorithm.
Instead, \mymodel provides a principled reframing that makes the decision structure underlying memory management explicit, clarifies the limitations of purely heuristic or myopic approaches, and establishes a foundation for the systematic design and analysis of uncertainty-aware memory systems for LLM-based agents.



\section{A Decision-Theoretic Framework for Agent Memory
Management}
\label{sec:approach}
\subsection{Narrative}

LLMs equipped with external memory must continuously decide which information to read, add, or delete as interactions unfold. 
These decisions shape future retrieval and downstream behavior, yet their utility is often delayed, indirect, and uncertain.
For example, consider a long-term assistant that gradually infers a user’s preferences or project context across many interactions. 
Prematurely deleting a seemingly irrelevant memory item, such as an early constraint or preference, may irreversibly impair future responses, while indiscriminate writing can pollute retrieval with redundant or spurious information whose harm may only manifest after many subsequent interactions, when corrective actions are no longer possible~\cite{lampinen2025latent}.

In practice, memory management is commonly implemented using heuristics such as recency rules, similarity thresholds, or fixed write policies~\cite{chhikara2025mem0,zhong2024memorybank}. 
While effective in specific settings, such heuristics lack a principled mechanism to reason about delayed consequences, trade-offs between competing operations, and uncertainty about future relevance. 
Consequently, these methods often fail in settings where memory utility is governed by long-term interaction patterns rather than immediate feedback, which are increasingly common in deployed LLM systems.

We argue that memory management is fundamentally a sequential decision problem under uncertainty, rather than a static bookkeeping or retrieval optimization task~\cite{powell2022sequential}. 
This perspective motivates a modular decision architecture in which candidate memory operations are proposed, evaluated, and arbitrated over time, explicitly accounting for delayed utility, irreversibility, and incomplete information.

\subsection{Basic Model}

We deliberately describe the memory management problem using the standard components of a sequential decision model, making explicit the information, decisions, and delayed consequences that heuristic implementations leave implicit.
This formulation specifies \emph{what information is available}, \emph{what decisions can be made}, and \emph{how decisions affect future states}, while remaining agnostic to any particular learning or optimization algorithm.

\subsubsection{State Variables}

At each decision step $t$, \mymodel observes a state $S_t \in \mathcal{S}$ that summarizes all information required to make a memory management decision.

Conceptually, $S_t$ may include three components: the current input or message, a compact summary of past interactions and auxiliary memory-related information such as metadata, usage statistics, or system constraints. 
The exact representation of $S_t$ is application-dependent; 
the only requirement is that it captures both the current context and the accessible memory upon which decisions are conditioned, highlighting that state design itself is a modeling choice that governs what aspects of history and memory are rendered decision-relevant.

\subsubsection{Decision Variables}

Given the state $S_t$, \mymodel selects a memory decision $A_t$.
The decision space is structured hierarchically to distinguish between information access and storage maintenance.

We represent the decision as a structured action tuple:
\[
A_t = (a_t^{\mathrm{read}}, a_t^{\mathrm{write}}) \in \mathcal{A}.
\]
Here, $a_t^{\mathrm{read}}$ governs information access, such as retrieval or context construction.
The storage maintenance component specifies how the memory contents are modified and is defined as
\[
a_t^{\mathrm{write}} = (a_t^{\mathrm{add}}, a_t^{\mathrm{delete}}),
\]
where $a_t^{\mathrm{add}}$ determines the addition of new memory items and $a_t^{\mathrm{delete}}$ specifies the removal of existing ones.
Memory updates are implicitly represented as a deletion followed by an addition, while selecting neither operation corresponds to a no-operation. \footnote{Both read and write actions are modeled as parameterized actions.
In practice, an action corresponds to selecting a tool together with its configuration parameters.
For notational simplicity, we omit explicit parameterization in the remainder of this section and treat each action as atomic, with all value and uncertainty estimates understood to operate over the underlying parameterized action space.}

Functionally, the read action typically precedes the write action, ensuring that memory modifications are conditioned on relevant retrieved information.
Feasibility constraints enforce internal consistency, such as preventing conflicting add and delete operations on the same memory item within a single decision step.
This structured formulation reflects the need to coordinate multiple, potentially interacting memory operations within a single decision step.

\subsubsection{Exogenous Information}

After a memory decision is made, \mymodel is exposed to new information that is not controlled by the memory mechanism. 
This exogenous information may include future user inputs, downstream task demands, or environmental signals that partially reveal the consequences of earlier memory decisions.

We denote this information by $W_{t+1}$. Importantly, $W_{t+1}$ is not known at the time the decision $A_t$ is made, which constitutes the primary source of uncertainty in memory management and precludes purely myopic or deterministic decision rules from reliably optimizing long-term memory utility.

\subsubsection{Transition Function}

\mymodel evolves as a result of both the chosen memory decision and the realized exogenous information:
\[
S_{t+1} = \mathcal{T}(S_t, A_t, W_{t+1}),
\]
where the transition function $\mathcal{T}$ captures modifications to the memory store, updates to summaries, and changes in retrieval behavior induced by new interactions.

\subsubsection{Objective Function}

The goal of the memory system is to support downstream performance over time while respecting operational constraints. We formalize this goal through a cumulative objective that aggregates utility across decision steps.

Let $C(S_t, A_t, W_{t+1})$ denote the contribution of a memory decision at step $t$, which may reflect downstream task performance, retrieval quality, or costs associated with memory operations.
Typical instantiations of $C$ may include proxies for task success (e.g., accuracy or user satisfaction), retrieval effectiveness (e.g., relevance or coverage), and explicit penalties for memory growth, write frequency, or deletion risk.
We consider both finite-horizon and discounted infinite-horizon settings; when long-term memory effects dominate, the latter is more appropriate. The objective is to select memory decisions that maximize expected cumulative utility:
\[
\max \;
\mathbb{E}
\left[
\sum_{t=0}^{\infty} \gamma^t \, C(S_t, A_t, W_{t+1})
\;\middle|\; S_0
\right].
\]
This objective makes explicit that memory decisions should be evaluated based on their long-term contribution to system behavior, rather than their immediate effect on a single interaction.

\subsection{Modeling Uncertainty}

A central challenge that distinguishes memory management from standard retrieval or compression is uncertainty about long-term utility.
The relevance of a memory item may only become apparent after many subsequent interactions, and observed outcomes are often noisy or indirect.

We model this uncertainty explicitly by associating each candidate memory operation with both an estimated utility and an uncertainty measure. 
The uncertainty considered here is primarily epistemic, arising from limited data, distributional shift, or incomplete observability of long-term downstream effects~\cite{zhou2022survey}.
In practice, such uncertainty may be approximated using Bayesian estimators, ensemble disagreement, bootstrapped value functions, or other predictive confidence measures
By treating uncertainty as a first-class signal, \mymodel can distinguish between actions that may be high-impact but poorly understood, and whose consequences may be revealed only after irreversible changes to the memory state.

\begin{figure}[t]
    \centering
    \includegraphics[width=0.5\textwidth]{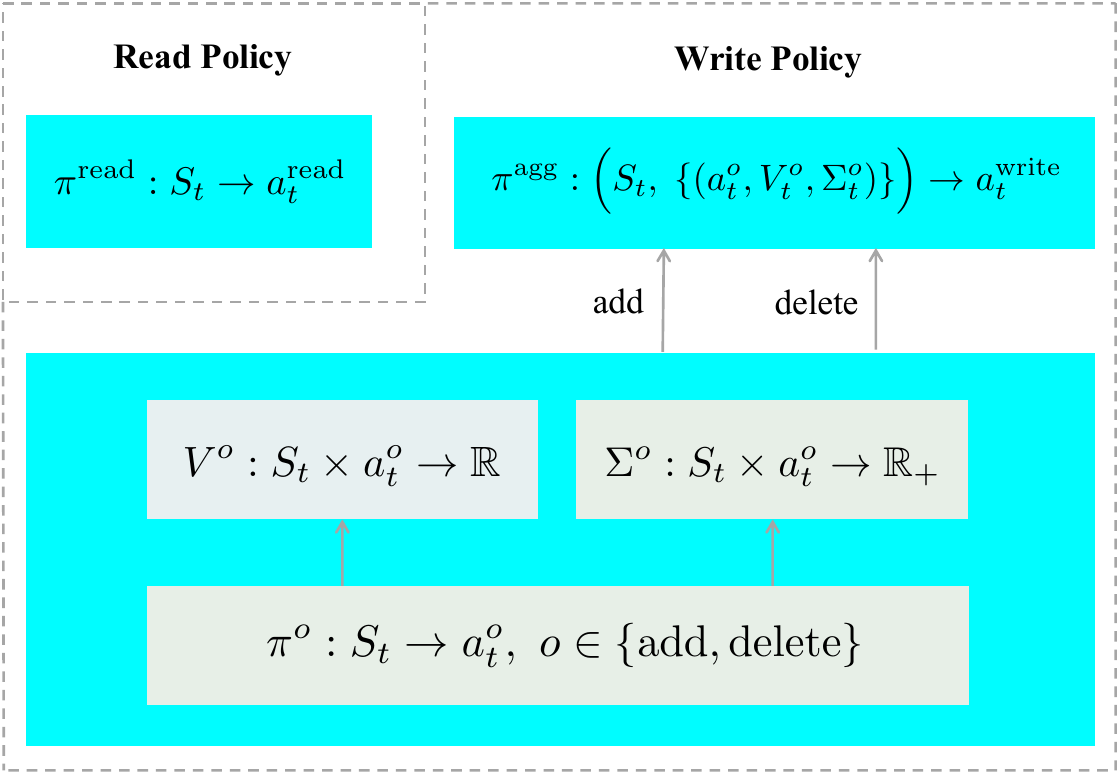} 
    \caption{
Overview of the hierarchical memory management policy. The read policy produces a retrieval action conditioned on the current state. Two write sub-policies (add, delete) independently propose candidate memory operations, each paired with a value estimate and an uncertainty score. The aggregate policy then arbitrates among these proposals to produce a coherent write action.
}

    \label{fig:policy}
\end{figure}

\subsection{Designing Policies}
Direct optimization over the full joint memory action space is typically intractable due to both combinatorial structure and severely delayed feedback, motivating a hierarchical decomposition rather than a monolithic policy.
We therefore introduce a hierarchical policy architecture that decomposes memory decisions into interpretable components (Figure~\ref{fig:policy}):
\[
\pi = (\pi^\mathrm{read}, \pi^\mathrm{add}, \pi^\mathrm{delete}, \pi^\mathrm{agg}).
\]
To address the challenges of delayed feedback and decision risk, this architecture is augmented with explicit value estimation and uncertainty quantification.
Specifically, each storage proposal is paired with a value function that predicts its long-term utility and an uncertainty estimator that gauges the confidence of that prediction.
These signals enable the aggregate policy to weigh expected benefits against the potential risks of irreversible memory modifications.
\textbf{Note that we adopt an asymmetric design:} these rigorous evaluations are applied exclusively to storage operations ($\mathcal{O}^\mathrm{write}$), whereas the read policy ($\pi^\mathrm{read}$) is modeled as a direct optimization of immediate relevance, reflecting the fact that retrieval actions are transient and do not permanently alter the underlying memory state.

\subsubsection{Read Policy}
The read policy $\pi^\mathrm{read}$ governs the information acquisition phase, operating as a functional prior to memory modification.
\[
\pi^\mathrm{read} : S_t \rightarrow a_t^\mathrm{read}.
\]
This policy determines the retrieval parameters (e.g., query generation or context filtering) to construct the effective context for subsequent operations. By optimizing $\pi_\mathrm{read}$, \mymodel ensures that the write policy is conditioned on the most relevant historical information.

\subsubsection{Write Policy}

The write policy governs the active maintenance of the memory store.
We denote the set of maintenance operations as $\mathcal{O}^\mathrm{write} = \{\mathrm{add}, \mathrm{delete}\}$.
Crucially, this module functions as a proposal mechanism rather than a direct executor.
It generates candidate operations along with their corresponding value and uncertainty estimates, which serve as inputs to the \textbf{aggregate policy}.
This separation ensures that potentially conflicting maintenance goals are arbitrated centrally before any irreversible modification is applied to the memory state.

\paragraph{Sub-policy.}
For each memory operation type $o \in \mathcal{O}^\mathrm{write}$, a dedicated sub-policy proposes a candidate action:
\[
\pi^o : S_t \rightarrow a_t^o.
\]
Each sub-policy addresses a localized decision problem, such as identifying salient information to add via $\pi^\mathrm{add}$ or detecting obsolete entries to remove via $\pi^\mathrm{delete}$, without explicitly reasoning about global trade-offs.
These localized proposals are generated conditionally on the current state $S_t$.

\paragraph{Value Estimation.}

To evaluate the proposed actions, each storage sub-policy is paired with a value function:
\[
V^o : S_t \times a_t^o \rightarrow \mathbb{R},
\qquad o \in \mathcal{O}^\mathrm{write}.
\]
Consistent with \mymodel objective, the value function estimates the expected discounted contribution of executing the corresponding proposal:
\[
V_t^o \approx
\mathbb{E}_\pi
\left[
\sum_{\tau=t}^{\infty} \gamma^{\tau-t} C(S_\tau, A_\tau, W_{\tau+1})
\;\middle|\;
S_t, a_t^o
\right].
\]
These operation-specific value functions act as critics that assess long-term impact (e.g., future retrieval utility) beyond immediate outcomes.

\paragraph{Uncertainty Quantification.}

To quantify the uncertainty associated with each proposed policy action, we introduce an uncertainty estimator for every operation type:
\[
\Sigma^o : S_t \times a_t^o \rightarrow \mathbb{R}_+.
\]
The resulting quantity
\[
\Sigma_t^o \approx
\mathtt{Uncertainty}\!\left(
a_t^o \mid S_t
\right)
\]
captures epistemic uncertainty in the policy's decision to propose action $a_t^o$, arising from limited evidence, distributional shift, or model mismatch. This signal is particularly important for high-stakes operations such as memory deletion ($\pi^\mathrm{delete}$), where unreliable policy decisions may lead to irreversible errors in the memory state.

\paragraph{Aggregate Policy.}

The aggregate policy serves as a central arbiter that integrates the retrieval decision and the localized write proposals into a coherent joint decision:
\[
\pi^\mathrm{agg}:
\Big(
S_t,\;
\{(a_t^o, V_t^o, \Sigma_t^o)\}_{o \in \mathcal{O}^\mathrm{write}}
\Big)
\rightarrow a_t^\mathrm{write}.
\]
Instead of greedily selecting the proposal with the highest estimated value in isolation, $\pi^\mathrm{agg}$ balances expected utility against uncertainty and feasibility constraints.
This arbitration may be instantiated via constrained optimization, risk-sensitive ranking, or threshold-based veto mechanisms that suppress high-uncertainty operations, particularly for irreversible actions such as deletion.
Formally, the aggregate policy selects a feasible joint action $a_t^\mathrm{write}$ that trades off expected contribution and risk of model error.

\subsection{Extensions}

The proposed framework naturally supports a range of extensions. 
State representations may be augmented with richer contextual signals or user attributes. Value and uncertainty estimators may be learned from data, approximated via simulation, or derived analytically. The arbitration mechanism may incorporate explicit constraints, memory budgets, or domain-specific rules.

Importantly, these extensions preserve the core sequential decision abstraction described above, enabling systematic comparison and incremental refinement of memory management mechanisms within a stable and interpretable decision-theoretic abstraction.

\section{Heuristic Memory vs. Decision-Theoretic Memory}
\label{sec:heuristic}

Most existing memory systems for LLMs rely on hand-designed heuristics~\cite{chhikara2025mem0,li2025memos,packer2023memgpt}. 
Common strategies include fixed rules for adding memories, deletion based on recency or similarity thresholds~\cite{zhong2024memorybank}, and update mechanisms driven by local criteria~\cite{fang2025lightmem,rasmussen2025zep}. 
While such heuristics are effective in static or narrowly defined settings, they lack a principled mechanism to reason about the long-term consequences of memory operations.

We contrast this heuristic paradigm with our proposed framework, which models memory management as a sequential decision problem under uncertainty.
This shift in perspective fundamentally alters the system design along five key dimensions.

\paragraph{From Static Rules to Sequential Optimization.}
Heuristic memory systems typically operate as static reaction rules, mapping a current trigger (e.g., a full context window) to a predetermined operation~\cite{li2025memos,xu2025mem}. 
Their behavior is justified empirically rather than derived from a formal objective.
In contrast, our framework defines memory management as an optimization of the cumulative objective $\sum \gamma^t C(S_t, A_t, W_{t+1})$. 
By formally defining state $S_t$ and action $A_t$, we treat memory not as a passive storage bucket, but as an active agent that must continuously choose between preserving information for future utility and conserving capacity to maintain performance~\cite{powell2022sequential}.

\paragraph{Temporal Reasoning and Delayed Utility.}
Heuristic approaches implicitly assume that the utility of a memory item can be assessed via immediate proxies, such as semantic similarity to the current query~\cite{zhong2024memorybank}. 
However, as noted in our narrative, the harm of prematurely deleting a constraint or polluting retrieval often manifests only after many subsequent interactions.
Our architecture explicitly addresses this temporal gap through the use of \textbf{Value Functions} ($V^o$). 
Instead of relying on immediate relevance, the system evaluates proposals based on their predicted impact on future states $S_\tau$ (where $\tau > t$), allowing the policy to retain currently irrelevant information that is crucial for long-term consistency.

\paragraph{Epistemic Uncertainty as a Decision Signal.}
A critical failure mode of heuristic systems is their deterministic treatment of ambiguous signals. 
For instance, a heuristic might delete a memory simply because it falls below a similarity threshold, ignoring the risk that the system's understanding of relevance might be flawed~\cite{packer2023memgpt}.
Our framework distinguishes between \emph{low utility} and \emph{unknown utility} by introducing explicit \textbf{Uncertainty Estimators} ($\Sigma^o$). 
By treating epistemic uncertainty as a first-class signal, the aggregate policy can implement conservative behaviors, such as inhibiting irreversible deletion operations when model confidence is low, thereby preventing catastrophic memory loss due to model error.

\paragraph{Structured Arbitration vs. Ad-Hoc Priorities.}
In heuristic systems, conflicts between operations (e.g., whether to update an existing entry or create a new one) are typically resolved through hard-coded priority lists or execution orders~\cite{chhikara2025mem0,wei2025mlp}.
Conversely, we employ a modular \textbf{Aggregate Policy} ($\pi^{\mathrm{agg}}$) that acts as a central arbiter. 
Rather than executing proposals largely in isolation, $\pi^{\mathrm{agg}}$ integrates the retrieval context, value estimates, and uncertainty scores from all sub-policies to select a joint action $A_t = (a_t^{\mathrm{read}}, a_t^{\mathrm{write}})$ that satisfies feasibility constraints. 
This ensures that memory maintenance is coherent and internally consistent.

\paragraph{Decomposition of Access and Maintenance.}
Finally, standard heuristics often conflate the logic for \emph{retrieving} information with the logic for \emph{storing} it.
Our model creates a strict separation between the \textbf{Read Policy} ($\pi^{\mathrm{read}}$), which optimizes for immediate informational needs, and the \textbf{Write Policy} ($\pi^{\mathrm{write}}$), which optimizes the long-term state of the memory store. 
This asymmetric design acknowledges that while retrieval is transient, storage operations induce persistent state changes that require more rigorous, risk-aware evaluation.

\section{Open Challenges and Research Directions}
\label{sec:discussion}

Having contrasted heuristic memory mechanisms with our decision-theoretic formulation, we now turn to the open challenges that remain.
While the proposed framework provides a rigorous structure for memory management by decomposing it into value estimation ($V$), uncertainty quantification ($\Sigma$), and policy arbitration ($\pi^{\mathrm{agg}}$), it does not prescribe unique implementations for these components.
Instead, it exposes a set of foundational questions that must be addressed to translate this perspective into robust and scalable systems.

\paragraph{The Credit Assignment Problem in Value Estimation.}
A central challenge lies in effectively learning or approximating the value function $V^o(S_t, a_t^o)$.
In our formulation, $V^o$ captures the discounted sum of future rewards $\sum \gamma^t C(\cdot)$.
However, in practice, the signal connecting a memory decision (e.g., deleting a constraint at $t=0$) to a downstream outcome (e.g., a hallucination at $t=100$) is extremely sparse and delayed~\cite{yu2025memagent,wang2025mem}.
Developing efficient methods to approximate $V^o$, perhaps through inverse reinforcement learning, trajectory synthesis, or offline critique, remains a critical hurdle for moving beyond myopic heuristics.

\paragraph{Inference Latency and Approximation.}
Explicitly computing value and uncertainty estimates for every candidate memory operation is computationally expensive.
A fully realized $\pi^{\mathrm{agg}}$ requires evaluating multiple potential writes and deletes at every step, which may be infeasible for high-throughput systems~\cite{bertsekas1997nonlinear}.
Future work must explore approximations that preserve the benefits of sequential reasoning while reducing cost, such as distilling the heavy aggregate policy into a lightweight value network or using heuristic gating to limit the number of proposals requiring rigorous evaluation~\cite{chhikara2025mem0,li2025memos}.

\paragraph{Calibrating Epistemic Uncertainty.}
Our framework relies on $\Sigma^o$ to veto high-risk actions during arbitration.
However, obtaining calibrated epistemic uncertainty from LLMs is notoriously difficult.
If $\Sigma^o$ underestimates the risk of a deletion, the system reverts to the brittle behavior of heuristics; if it overestimates risk, the memory becomes stagnant~\cite{powell2022sequential}.
Determining which forms of uncertainty (e.g., ensemble disagreement vs. semantic entropy) provide the most reliable signal for memory arbitration is an essential open question~\cite{zhong2024memorybank}.

\paragraph{Joint Optimization of Read and Write Policies.}
While we model $\pi^{\mathrm{read}}$ and $\pi^{\mathrm{write}}$ as distinct components to separate concerns, they are functionally coupled.
The quality of a write decision depends entirely on the information retrieved by the read policy; conversely, the efficacy of the read policy depends on the state maintenance performed by the write policy~\cite{lample2019large}.
Optimization is difficult because $\pi^{\mathrm{read}}$ modifies the immediate context (transient) while $\pi^{\mathrm{write}}$ modifies the storage (persistent).
Jointly optimizing these policies without destabilizing the learning process remains a complex control problem~\cite{yan2025memory,pritzel2017neural}.

\paragraph{State Representation and Multi-Context Extension.}
Our current definition of state $S_t$ focuses on a single interaction history.
Extending this to settings involving multiple users or shared memory pools requires augmenting $S_t$ to capture distribution shifts and interference effects~\cite{packer2023memgpt,xu2025mem}.
In such scenarios, the objective function must also account for privacy and fairness, complicating the definition of the reward signal $C(S_t, A_t, W_{t+1})$.

\paragraph{Policy-Oriented Evaluation.}
Finally, validating this framework requires new evaluation protocols.
Existing benchmarks often measure downstream task performance directly, conflating the quality of memory decisions with the reasoning capability of the LLM~\cite{ai2025memorybench,zhangConversationalAgentsRAG2025}.
To rigorously test the proposed architecture, we need diagnostic benchmarks that specifically isolate the accuracy of $V^o$ (predictive validity) and the calibration of $\Sigma^o$, rather than just the final output of the system~\cite{hu2025memory}.

In summary, we view decision-theoretic memory not as a turnkey solution, but as a unifying roadmap.
By making the sequential and uncertain nature of memory decisions explicit, this view clarifies \emph{what} must be solved, estimating long-term value and managing risk, even if the specific algorithms for doing so remain an active frontier of research.

\section{Related Work}
\label{sec:realted}

Our work sits at the intersection of memory systems, language modeling, and sequential decision-making.
Rather than proposing a specific architecture or a singular optimization algorithm, we aim to provide a unifying decision-theoretic perspective on the memory lifecycle.
Below, we situate our contribution within the broader literature\footnote{\href{https://github.com/TeleAI-UAGI/awesome-agent-memory}{Awesome-Agent-Memory (GitHub)}}.

\paragraph{Memory-Augmented Models.}
Early efforts to extend neural networks with external memory focused on differentiable architectures capable of algorithmic manipulation.
Pioneering works such as Neural Turing Machines~\cite{graves2014neural} and Differentiable Neural Computers~\cite{graves2016hybrid} introduced mechanisms for reading from and writing to memory matrices.
Later approaches scaled these ideas using content-addressable key-value stores~\cite{sukhbaatar2015end,santoro2016meta} or sparse retrieval mechanisms~\cite{lample2019large,pritzel2017neural}.
While foundational, these works primarily address the \emph{architecture} of memory access (how to represent and query memory) rather than the \emph{management policy} (deciding what to retain or discard over long horizons), which remains largely implicit or fixed in these models.

\paragraph{Heuristic Memory Management in LLMs.}
In modern LLM deployments, memory management is predominantly driven by static heuristics (e.g. TeleMem). 
Although the underlying representations vary widely, ranging from raw text~\cite{chhikara2025mem0,li2025memos} and knowledge graphs~\cite{rasmussen2025zep,xu2025mem} to parametric encodings~\cite{liu2024generation,wei2025mlp}, the policies governing these memories are remarkably similar.
Common strategies include sliding windows, periodic summarization~\cite{packer2023memgpt,fang2025lightmem}, or pruning based on simple similarity thresholds~\cite{zhong2024memorybank}.
While effective for specific use cases, these hand-crafted rules are rigid: they do not adapt to shifting user goals and offer no formal mechanism to weigh the trade-offs between storage costs and future utility.
Consequently, decisions about forgetting or updating are often decoupled from their downstream consequences.

\paragraph{Learning Memory Operations via RL}
Recognizing the limitations of heuristics, recent research has begun to formulate memory operations as learnable actions optimized via reinforcement learning (RL)~\cite{yu2025memagent,zheng2025goal}.
Specific methods have been proposed to optimize distinct sub-problems: for instance, Memento~\cite{yan2025memory} trains a policy to filter retrieval to reduce noise, while Mem-$\alpha$~\cite{wang2025mem} focuses on selective writing to prevent memory saturation.
While these works demonstrate that memory operations \emph{can} be learned, they typically treat the decision process as a black-box optimization problem or focus on isolated components (e.g., only reading or only writing).
Our work complements this direction by providing the explicit theoretical scaffolding, including the definition of the state space, decision variables, and information flow, which is necessary to rigorously model, analyze, and unify these disparate learning-based approaches.

\paragraph{Sequential Decision Perspectives.}
Finally, our framework \mymodel draws on the rich tradition of sequential decision-making under uncertainty~\cite{bellman1966dynamic,powell2022sequential}.
This field emphasizes that decisions made today (e.g., storing a user preference) change the state of the system for all future time steps, often with delayed and stochastic feedback~\cite{puterman2014markov,bertsekas1997nonlinear}.
While these principles are standard in control theory and operations research, their application to LLM memory is distinct due to the unstructured, high-dimensional nature of language data.
We bridge this gap by mapping the vague notion of ``memory management'' onto formal decision-theoretic components, offering a principled language to reason about uncertainty and long-term value in LLMs.

\section{Conclusion}
\label{sec:conclusion}

In this work, we have argued that memory management for LLMs is best understood as a sequential decision problem under uncertainty, rather than as a collection of hand-designed heuristics.
By formalizing memory management as a hierarchy of immediate access and long-term maintenance, we exposed the critical need to balance current relevance with future utility.
Our proposed framework explicitly addresses this challenge by introducing value estimation and uncertainty quantification as central signals for arbitrating memory operations.
This perspective not only clarifies the implicit trade-offs within existing heuristic approaches but also provides a rigorous structure for reasoning about delayed consequences and risk.
As LLMs evolve into long-lived agents, we believe that moving from static rules to decision-theoretic control, governed by explicit objectives and uncertainty, is essential for building robust and adaptive memory systems.


\bibliography{main}



\end{document}